\documentclass{article}

\usepackage[preprint]{neurips_2026}

\usepackage[utf8]{inputenc} 
\usepackage[T1]{fontenc}    
\usepackage{hyperref}       
\usepackage{url}            
\usepackage{booktabs}       
\usepackage{amsfonts}       
\usepackage{nicefrac}       
\usepackage{microtype}      
\usepackage{xcolor}         

\usepackage{amsmath}
\usepackage{amsthm}
\usepackage{bm}
\usepackage{bbm}
\usepackage{graphicx}
\usepackage{subcaption}

\usepackage{algorithm}
\usepackage[noend]{algpseudocode}

\bibpunct{(}{)}{;}{a}{,}{,}

\DeclareMathOperator*{\argmax}{arg\,max}
\def\R{\mathbb{R}}
\def\X{\mathcal{X}}

\newcommand{\set}[1]{\left\{#1\right\}}

\newcommand{\cc}[1]{\left[#1\right]}
\DeclareMathOperator*{\E}{\mathbb{E}}

\title{Elicitation-Augmented Bayesian Optimization}

\author{
  Alvar Haltia \\
  Aalto University \\
  \texttt{alvar.haltia@aalto.fi} \\
  \And
  Ville Hyvönen \\
  University of Helsinki \\
  \texttt{ville.o.hyvonen@helsinki.fi} \\
  \And
  Samuel Kaski \\
  Aalto University, University of Manchester  \\
  \texttt{samuel.kaski@aalto.fi} \\
}

\begin{document}

\maketitle

\begin{abstract}

Human-in-the-loop Bayesian optimization (HITL BO) methods utilize human expertise to improve the sample-efficiency of BO. Most HITL BO methods assume that a domain expert can quantify their knowledge, for instance by pinpointing query locations or specifying their prior beliefs about the location of the maximum as a probability distribution. However, since human expertise is often tacit and cannot be explicitly quantified, we consider a setting where domain knowledge of an expert is elicited via pairwise comparisons of designs. We interpret the expert's pairwise judgements as noisy evidence about the values of the observable objective function and develop a principled method for combining the information obtained via direct observations and pairwise queries. Specifically, we derive a cost-aware value-of-information acquisition function that balances direct observations against pairwise queries. The proposed method approaches the convex hull of the trajectories of the individual information sources: when pairwise queries are cheap it substantially improves sample-efficiency over observation-only BO, and when pairwise queries are costly or noisy, it recovers the performance of standard BO by relying on direct observations alone.

\end{abstract}

\section{Introduction}

Bayesian optimization (BO) is a sample-efficient framework for optimizing expensive-to-evaluate black-box functions \citep{shahriari2016taking,frazier2018tutorial,garnett2023bayesian}. In many real-world applications---such as aerodynamic and material design \citep[e.g.,][]{adachi2021high,cisse2024hypbo}--- human experts have domain knowledge that is cheaper to elicit compared to directly evaluating the objective function. Hence, there is a growing body of \emph{human-in-the-loop} (HITL) BO methods \citep[e.g.,][]{arun2022human,mikkola2023multi,hvarfner2024general,xu2024principled} that utilize this domain knowledge to improve the cost-efficiency of BO.

Most of the existing HITL BO methods assume that a domain expert can quantify their knowledge, for instance by pinpointing query locations \citep{arun2022human,gupta2023bomuse,khoshvishkaie2023cooperative} or by specifying their prior beliefs about the location of the maximum as a probability distribution \citep{hvarfnerpi2022bo,hvarfner2024general,cisse2024hypbo}. However, human expertise is often tacit and qualitative \citep{kahneman1979interpretation}. Humans rank candidates more reliably than they attach calibrated numbers \citep{shah2014better}, and, consequently, pairwise comparisons are a natural type of query for knowledge elicitation.

We consider both single- and multi-objective Bayesian optimization. To extend knowledge elicitation via pairwise comparisons to the multi-objective case, we assume that the utility function is elicited or specified prior to the optimization process. When the utility function is known, the expert's pairwise judgements can be interpreted unambiguously: they constitute noisy evidence about the objective function itself. Direct function evaluations and pairwise comparisons then inform a single shared surrogate, and single-objective and multi-objective optimization are handled uniformly.

We develop a principled Gaussian process (GP)-based method for combining data from direct function evaluations and pairwise comparisons. Specifically, we use a mixed-likelihood sparse variational GP that couples a Gaussian observation likelihood and a probit comparison likelihood, yielding a decomposable evidence lower bound (ELBO) with a closed-form regression term and a Gauss--Hermite probit term. Because both information sources are used to update the same posterior, the optimization reduces to a sequential source-selection problem: at each iteration, a possible action is either a direct evaluation or a pairwise comparison. Following the adaptive-elicitation framework by \citet{ungredda2023elicit}, we score each candidate action by its expected one-step utility gain per unit cost---a cost-normalized value-of-information rule. For evaluations, this reduces to the knowledge gradient \citep{frazier2009knowledge}. For pairwise comparisons, we derive a closed-form fantasy posterior update under the probit comparison likelihood for a multi-output objective under a known utility \citep{wu2026knowledge}. Both acquisition functions are optimized jointly with candidate inputs via the one-shot reformulation \citep{balandat2020botorch}.

We evaluate the proposed method, called Elicitation-Augmented Bayesian Optimization (\textsc{EA-BO}), on standard single- and multi-objective benchmarks. Across cost regimes, \textsc{EA-BO} approaches the convex hull of the two single-source trajectories: it allocates the budget to the more cost-effective source at each phase of the optimization, outperforming evaluation-only BO when pairwise queries are cheap and recovering the performance of standard BO when pairwise queries are costly or noisy. \textsc{EA-BO} also substantially outperforms the earlier HITL BO method CoExBO \citep{adachi2024looping} that, like our work, assumes that the expert is able to express their domain knowledge by performing pairwise comparisons.

To summarize, our main contributions are: \textbf{(i)} We derive a mixed-likelihood sparse variational GP for updating the posterior of the model combining direct objective function evaluations with pairwise queries. \textbf{(ii)} We derive a cost-aware value-of-information acquisition rule with a closed-form posterior mean under the probit comparison likelihood, extending the pairwise knowledge gradient of \citet{wu2026knowledge} to a multi-output objective under a known utility. \textbf{(iii)} The experiments on standard benchmark functions (Section~\ref{sec:experiments}) indicate that the proposed method traces the Pareto-optimal convex hull over the choice of information sources, thus substantially improving the cost-efficiency of BO in the regime where pairwise comparisons are cheaper than evaluations.

\section{Related Work}
\label{sec:related}

\textbf{Human-in-the-loop BO.} Earlier HITL BO methods typically assume that the expert can quantify their domain knowledge explicitly. An approach inspired by classic Bayesian inference requires that the expert specifies their prior beliefs about the location of the maximum of the objective function as a probability distribution \citep{ramachandran2020incorporating,souza2021bayesian,hvarfnerpi2022bo,hvarfner2024general,cisse2024hypbo,guayhottin2025robust}. In the human-AI collaborative approach the expert pinpoints potential locations for the next objective function evaluation \citep{arun2022human,gupta2023bomuse,khoshvishkaie2023cooperative}. In contrast to these earlier methods, we assume that the domain knowledge of the expert is tacit, and perform the knowledge elicitation implicitly via pairwise queries. Another method that uses pairwise queries is CoExBO \citep{adachi2024looping}. The main differences to our method are that they use pairwise queries to directly select the location of the next function evaluation, select the locations of the pairwise queries at random, an do not account for the query costs. Another thematically similar work is \citet{xu2024principled}, who also utilise a human expert with tacit knowledge, but assumes the expert accepts or rejects designs. The work is not directly comparable due to differing assumptions on the expert. 

\textbf{Preferential BO and multi-objective BO.} Preferential BO \citep{chu2005preference,gonzalez2017preferential,siivola2021preferential,mikkola2020projective,benavoli2021preferential,takeno2023practical,astudillo2023qeubo,xu2024principled_pbo} and interactive multi-objective BO \citep{astudillo2020multiattribute,ungredda2021onestep,lin2022preference,ozaki2024multi} also utilize pairwise queries. In contrast to our method, which uses pairwise queries as an additional information source about the values of the observable objective function, preferential BO and multi-objective BO use pairwise queries to optimize the latent preference or utility function, respectively. Observe also that in interactive multi-objective optimization the expert compares two points in the objective space $\mathbb{R}^m$, whereas we compare two points in the design space $\mathcal{X}$. Composite-function BO \citep{astudillo2019composite} similarly exploits a known utility structure but considers only direct evaluations of the objective.

\textbf{Mixed likelihood methods.} The recent work by \citet{wu2025mixed} also considers updating a GP-based surrogate model using two different information sources. However, they combine pairwise queries and data on Likert scale, whereas we combine pairwise queries with direct function evaluations. \citet{zhang2020binary} use binary auxiliary functions as lower-fidelity information sources, but they consider a setting with two correlated surrogate models.

\section{Preliminaries}
\label{sec:problem_setup}

We consider the task of optimizing a vector-valued \emph{objective function} 
\[
f: \X \rightarrow \R^m
\]
over a bounded \emph{design space} $\X \subset \R^d$. We assume that $f$ is an expensive-to-evaluate black box function, i.e., we do not know its functional form or gradients. Given a known \emph{utility function} $U: \R^m \rightarrow \R$, the goal of the optimization process is to find
\begin{equation*}
    x^* = \argmax_{x \in \X}\, U(f(x)),
\end{equation*}
i.e., the design with the highest expected utility. Standard single-objective BO is a special case with $m=1$ and $U(y) = y$. When $m > 1$, utility function $U$ encodes the decision maker's priorities over the Pareto front---for instance, a weighted sum or Chebyshev scalarization. We consider two information sources for learning the values of $f$ : direct function evaluations and pairwise queries.

\textbf{Direct observations.} The experimenter selects $x \in \X$, pays a fixed cost $c_{\mathrm{eval}} \in \R_+$, and receives
\begin{equation*}
    y = f(x) + \varepsilon_{\mathrm{eval}}, \qquad \varepsilon_{\mathrm{eval}} \sim N(0, \sigma_{\mathrm{eval}}),
\end{equation*}
where $\sigma_{\mathrm{eval}} = \mathrm{diag}(\sigma_{\mathrm{eval},1}^2, \dots, \sigma_{\mathrm{eval},m}^2)$ is the diagonal covariance matrix defining the observation noise.

\textbf{Pairwise queries.} We assume that there is a data generating process yielding comparisons of two competing designs $x,x' \in \mathcal{X}$, and that their ranking is governed by $U \circ f$ (with additional noise); a human expert is the most plausible data source. Formally, the experimenter selects a tuple $(x, x') \in \X^2$, pays a cost $c_{\mathrm{comp}} \in \R_+$, and receives a binary judgement $d \in \set{0,1}$, where $d = 1$ indicates that $x$ is ranked higher than $x'$. Using the probit link function, the likelihood can be written as 
\begin{equation*}
    p(d = 1 \mid x, x', f) = \Phi\!\left(\frac{U(f(x)) - U(f(x'))}{\sqrt{2}\, \sigma_{\mathrm{comp}}}\right),
\end{equation*}
where $\Phi$ is the CDF of $N(0,1)$ and $\sigma_{\mathrm{comp}} > 0$ is the noise parameter of the pairwise comparisons.

\textbf{Notation.} We denote the direct function evaluations by $\mathcal{D}_{\mathrm{eval}} = \set{(x_i, y_i)}_{i=1}^{n_{\mathrm{eval}}}$, the responses to pairwise queries by $\mathcal{D}_{\mathrm{comp}} = \set{((x_j, x_j'), d_j)}_{j=1}^{n_\mathrm{comp}}$, and the set of all observations by $\mathcal{D} = \{\mathcal{D}_{\mathrm{eval}}, \mathcal{D}_{\mathrm{comp}}\}$.

\section{Surrogate model}
\label{sec:surrogate}

In this section, we first describe a GP-based surrogate model of the objective function we use for modelling the data from the task described in Section~\ref{sec:problem_setup}, and then derive a variational inference algorithm for updating the posterior of the model.

\subsection{GP Model}

We place a zero-mean Gaussian process prior on $f$ with independent coordinates:
\begin{equation*}
    f_j \sim \mathcal{GP}(0, k_j), \qquad j = 1, \dots, m,
\end{equation*}
where each $k_j$ is an ARD squared-exponential kernel with lengthscales $\ell_{j,p}$ and output-scale $\sigma_j^2$. We place Gamma hyperpriors on $\ell_{j,p}$ and $\sigma_j^2$, and write $\theta$ for the collection of all kernel hyperparameters and noise variances.

Combining this prior with the likelihoods of Section~\ref{sec:problem_setup}, the full generative model is
\begin{align*}
    \theta &\sim p(\theta), \\
    f_j &\sim \mathcal{GP}(0, k_j(\cdot, \cdot; \theta)) \qquad \text{independently for } j = 1, \dots, m, \\
    y_i \mid x_i, f &\sim N(f(x_i), \sigma_{\mathrm{eval}}), \\
    d_j \mid (x_j, x'_j), f &\sim \mathrm{Bern}\!\left(\Phi\!\left(\tfrac{U(f(x_j)) - U(f(x'_j))}{\sqrt{2}\,\sigma_{\mathrm{comp}}}\right)\right),
\end{align*}
and the posterior follows by Bayes' rule,
\begin{equation*}
    p(f \mid \mathcal{D}, \theta) \propto p(f \mid \theta) \prod_{i=1}^{n_\mathrm{eval}} p(y_i \mid x_i, f) \prod_{j=1}^{n_\mathrm{comp}} p(d_j \mid x_j, x'_j, f).
\end{equation*}
Note that both likelihood terms share $f$. We denote the posterior moments by $\mu_\mathcal{D}(x) := \E\cc{f(x) \mid \mathcal{D}}$ and $K_\mathcal{D}(x, x') := \mathrm{Cov}\cc{f(x), f(x') \mid \mathcal{D}}$, dropping the subscript when the dataset is unambiguous.

\subsection{Posterior inference}
\label{sec:vi}

The posterior $p(f \mid \mathcal{D}, \theta)$ has no closed form solution, and thus we have to approximate it. Sparse variational inference is particularly well-suited to this setting, since the Sparse Variational Gaussian Process (SVGP) evidence lower bound (ELBO) factorizes over the observations and comparisons, enabling stochastic optimization that scales to large datasets, and---as we show below---the dual-likelihood structure of our model yields separate, tractable contributions to the bound. We follow the standard sparse variational GP construction \citep{titsias2009variational,hensman2013gaussian,hensman2015scalable}, introducing $M$ inducing locations $Z \subset \X$ with inducing values $u = f(Z)$ and parametrizing the variational posterior as
\begin{equation*}
    q(f, u) = p(f \mid u)\, q(u), \qquad q(u) = N(u \mid m_u, S_u),
\end{equation*}
where $S_u = L_u L_u^\top$ is a Cholesky parametrization. We denote the moments of the variational posterior $q$ by $\mu(\cdot)$ and $K(\cdot, \cdot)$.

The dual-likelihood structure of our dataset yields a decomposable ELBO with separate observation and query contributions \citep{wu2025mixed,moreno2018heterogeneous}:
\begin{equation*}
\begin{split}
    \mathcal{L}(q, \theta) & = \sum_{i=1}^{n_\mathrm{eval}} \E_{q(f(x_i))}\!\cc{\log p(y_i \mid x_i, f)} \\
    &+ \sum_{j=1}^{n_\mathrm{comp}} \E_{q(f(x_j), f(x'_j))}\!\cc{\log p(d_j \mid x_j, x'_j, f)} - \mathrm{KL}\!\left(q(u) \,\|\, p(u \mid \theta)\right).    
\end{split}
\end{equation*}
The evaluation term and the KL admit standard Gaussian closed forms (Appendix~\ref{app:elbo_details}). The pairwise comparison term is an expectation of $\log \Phi(\cdot)$ under the joint marginal $q(f(x_j), f(x'_j))$ and requires more care. Because the integrand depends on $(f(x_j), f(x'_j))$ only through the scalar utility difference $\Delta_j = U(f(x_j)) - U(f(x'_j))$, the expectation reduces to a one-dimensional integral against the distribution of $\Delta_j$ under $q$. For linear $U$, $\Delta_j$ is exactly Gaussian and the expectation can be computed by one-dimensional Gauss--Hermite quadrature. The nonlinear case is handled by moment matching and deferred to Appendix~\ref{app:nonlinear}.

We optimize $\mathcal{L}$ by gradient ascent over the variational parameters $(Z, m_u, L_u)$, the kernel hyperparameters $\theta$, and the noises $\sigma_{\mathrm{comp}}, \sigma_{\mathrm{eval}}$ jointly with Adam. The hyperpriors on $\theta$ enter as an additive $\log p(\theta)$ regularizer.

\section{Acquisition function}
\label{sec:method}

In this section, we first define the decision problem, and then derive the one-step utility gain for this problem under the surrogate model described in Section~\ref{sec:surrogate}. The acquisition function is a cost-weighted sum over the contributions of the two possible actions: a direct function evaluation or a pairwise query.

\subsection{Decision problem}
\label{sec:decision_problem}

The optimization proceeds under a fixed total budget $\Lambda \in \R_+$. We select the recommended design after the budget is exhausted as 
\[
\hat{x} = \argmax_{x \in \X}\, \E_f\cc{U(f(x)) \mid \mathcal{D}},
\]
and hence the utility of the dataset $\mathcal{D}$ as $u(\mathcal{D}) = \max_{x \in \X}\, \E_f\cc{U(f(x)) \,\big|\, \mathcal{D}}$, where the expectation is over the uncertainty of the function approximation. Since the maximum is taken over the whole design space $\X$, the acquisition function we derive belongs to the knowledge gradient-family.

At each iteration, the experimenter either directly evaluates the objective function at $x \in \X$ or performs a pairwise query for the pair $(x,x') \in \X^2$, i.e., selects an action from the action space $\mathcal{A} = \X \,\cup\, \X^2$,
and pays the cost $c_{\mathrm{eval}}$ or $c_\mathrm{comp}$ depending on the type of action chosen. 
This decision problem is an instance of cost-aware multi-information-source optimization \citep{swersky2013multi,kandasamy2017multifidelity,poloczek2017multi}. Exact optimization of the full action sequence is intractable; we adopt the standard myopic one-step lookahead \citep{frazier2018tutorial}.

After selecting action $a \in \mathcal{A}$ and receiving result $r$ ($r \in \mathbb{R}^m$ or $r \in \{0,1\}$ depending on whether the action was a direct observation or a pairwise comparison, respectively) the updated dataset is $\mathcal{D}' = \mathcal{D} \cup \set{(a, r)}$. The \emph{value of information} (VoI) 
\begin{equation*}
    \alpha(a; \mathcal{D}) = \E_r\cc{u(\mathcal{D}') \,\big|\, a, \mathcal{D}} - u(\mathcal{D})
\end{equation*}
of action $a$ is the expected one-step gain obtained by performing it.

We select the action maximizing cost-normalized VoI \citep{poloczek2017multi,ungredda2023elicit}
\[
a^* \in \argmax_{a \in \mathcal{A}}\; \frac{\alpha(a;\mathcal{D})}{c(a)},
\]
where
\[
\alpha(a;\mathcal{D}) = \begin{cases}
\alpha_\mathrm{eval}(a;\mathcal{D}), \,\, &\mathrm{when} \,\, a \in \mathcal{X} \\
\alpha_\mathrm{comp}(a;\mathcal{D}), &\mathrm{when} \,\, a \in \mathcal{X}^2; \\
\end{cases}
\quad
c(a) = \begin{cases}
c_\mathrm{eval}, \,\, &\mathrm{when} \,\, a \in \mathcal{X} \\
c_\mathrm{comp},  &\mathrm{when} \,\, a \in \mathcal{X}^2. \\
\end{cases}
\]
Here $\alpha(a;\mathcal{D})$ and $c(a)$ denote the VoI and cost of action $a$---that is, $\alpha_{\mathrm{eval}}(x;\mathcal{D})$ at cost $c_{\mathrm{eval}}$ when $a \in \X$, and $\alpha_{\mathrm{comp}}((x,x');\mathcal{D})$ at cost $c_{\mathrm{comp}}$ when $a \in \X^2$; these terms are derived below in Sections~\ref{sec:VoI_observations} and \ref{sec:VoI_pairwise_comparisons}.

We select cost normalization over cost subtraction since under a fixed budget, cost normalization maximizes the rate of expected utility gain per unit budget spent, whereas cost subtraction can favor expensive actions whose absolute VoI is high but whose per-unit return is poor \citep{poloczek2017multi}.

Algorithm~\ref{alg:both} summarizes a single iteration\footnote{Per-iteration cost of \textsc{EA-BO} is dominated by the ELBO fit, which scales as $O((n_\mathrm{eval} + n_\mathrm{comp}) M^2 + M^3)$, and by the one-shot acquisition optimization which is linear in the number of fantasy points $K$ and gradient steps.} of the proposed algorithm \textsc{EA-BO}. At each step we fit the variational posterior, optimize both VoIs to obtain their respective maxima $\alpha_\mathrm{eval}^\star$ and $\alpha_{\mathrm{comp}}^\star$, and then select the action with highest cost-normalized VoI. The optimization loop terminates when the remaining budget falls below $\min(c_{\mathrm{eval}}, c_{\mathrm{comp}})$, and the final recommendation is $\hat{x} = \argmax_{x \in \X}\, \E_f\cc{U(f(x)) \mid \mathcal{D}_\mathrm{final}}$.

\begin{algorithm}[tb!]
\caption{Elicitation-Augmented Bayesian Optimization (\textsc{EA-BO}).}
\label{alg:both}
\begin{algorithmic}[1]
\State \textbf{Input:} design space $\X \subset \R^d$, utility $U$, budget $\Lambda$, costs $c_{\mathrm{eval}}, c_{\mathrm{comp}}$, initial design $\mathcal{D}_0 = \emptyset$, hyperparameter priors.
\State \textbf{Output:} recommended design $\hat{x} \in \X$.
\State Fit $q_0$ on $\mathcal{D}_0$ (ELBO of Section~\ref{sec:vi}). Set $t \gets 0$, $\Lambda_t \gets \Lambda$.
\While{$\Lambda_t \geq \min(c_{\mathrm{eval}}, c_{\mathrm{comp}})$}
    \State $x_t^* \gets \argmax_x \widehat{\alpha}_\mathrm{eval}(x; \mathcal{D}_t)$.
    \State $(x_a, x_b)_t^* \gets \argmax_{x_a, x_b} \widehat{\alpha}_{\mathrm{comp}}((x_a, x_b); \mathcal{D}_t)$.
    \State Restrict to actions affordable under $\Lambda_t$.
    \State $a_t^* \gets \argmax_i\, \widehat{\alpha}_i^\star / c_i$ over the affordable actions.
    \If{$a_t^* = x_t^* \in \mathcal{X}$, i.e. an evaluation}
        \State Observe $y_t \sim N(f(x_t^*), \Sigma_{\mathrm{eval}})$. Set $\mathcal{D}_{t+1} \gets \mathcal{D}_t \cup \set{(x_t^*, y_t)}$, $\Lambda_{t+1} \gets \Lambda_t - c_{\mathrm{eval}}$.
    \Else
        \State Query expert on $(x_a, x_b)_t^*$. Set $\mathcal{D}_{t+1} \gets \mathcal{D}_t \cup \set{((x_a, x_b)_t^*, d_t)}$, $\Lambda_{t+1} \gets \Lambda_t - c_{\mathrm{comp}}$.
    \EndIf
    \State Refit $q_{t+1}$ on $\mathcal{D}_{t+1}$, warm-started from $q_t$.
    \State $t \gets t + 1$.
\EndWhile
\State \textbf{return} $\hat{x} = \argmax_{x \in \X} \E\!\cc{U(f_{q_\tau}(x))}$.
\end{algorithmic}
\end{algorithm}

\subsection{Value-of-information for a direct observation}
\label{sec:VoI_observations}

The VoI for a function evaluation, denoted by $\alpha_{\mathrm{eval}}$, is the expected one-step gain in dataset utility after observing $f$ at $x \in \mathcal{X}$. Because the observation likelihood is Gaussian, the posterior mean $\mu_{\mathcal{D}'}$ is available in a closed form, and for linear $U$, $\alpha_{\mathrm{eval}}$ reduces to knowledge gradient \citep{frazier2009knowledge,wu2016parallel}. For nonlinear $U$, see Appendix~\ref{app:dataset_utility}. We adopt the one-shot reformulation by \citet{balandat2020botorch} and \citet{wu2020practical}, introducing $K$ fantasy evaluation points $x'_1, \dots, x'_K \in \X$ and jointly optimizing
\begin{equation*}
    \widehat{\alpha}_{\mathrm{eval}}(x, x'_1, \dots, x'_K; \mathcal{D}) = \E_y\!\cc{\max_{k \in \set{1,\dots,K}} \E_{\mathcal{D}'}\!\left[U(f(x'_k))\right]} - u(\mathcal{D}),
\end{equation*}
where the expectation is approximated by reparametrized Monte Carlo samples of $y$.

\subsection{Value-of-information for pairwise comparisons}
\label{sec:VoI_pairwise_comparisons}

The VoI of a pairwise comparison measures the expected one-step gain from querying the pair $(x_a, x_b)$. The outcome $d \in \set{0,1}$ has posterior predictive distribution
\begin{equation*}
    p(d = 1 \mid x_a, x_b, \mathcal{D}) = \Phi\!\left(\frac{\mu_\Delta}{\sqrt{\sigma_\Delta^2 + 2\sigma_{\mathrm{comp}}^2}}\right),
\end{equation*}
where $\mu_\Delta$ and $\sigma_\Delta^2$ are the mean and variance of $\Delta = U(f(x_a)) - U(f(x_b))$ under $q$. The comparison VoI is a discrete two-term expectation over the binary outcome,
\begin{equation*}
    \alpha_{\mathrm{comp}}((x_a, x_b); \mathcal{D}) = \sum_{d \in \set{0,1}} p(d \mid x_a, x_b, \mathcal{D})\, \max_{x \in \X} \E_{\mathcal{D}'}\!\cc{U(f(x))} - u(\mathcal{D}),
\end{equation*}
and thus can be evaluated exactly without Monte Carlo approximation.

The key quantity is the posterior mean after observing $d$. For linear $U$, $\Delta$ is Gaussian since it is a linear functional of the jointly Gaussian $(f(x_a), f(x_b))$ under $q$. For nonlinear $U$, see Appendix~\ref{app:fantasy_updates}. \citet{wu2026knowledge} derive a closed-form conditional mean for probit comparisons over a scalar $f$. We extend their result to a multi-output objective under a known utility $U$. In the case $d=1$, the fantasy posterior mean is
\begin{equation*}
    \mu_{\mathcal{D} \cup \set{((x_a, x_b), 1)}}(x) = \mu(x) + \frac{\phi(\tau)}{\Phi(\tau)} \cdot \frac{\mathrm{Cov}_{\mathrm{comp}}\cc{f(x), \Delta}}{\nu},
\end{equation*}
where $\nu := \sqrt{\sigma_\Delta^2 + 2\sigma_{\mathrm{comp}}^2}$ and $\tau := \mu_\Delta / \nu$. The result in case $d = 0$ is obtained by replacing $\tau$ with $-\tau$. The proof is deferred to Appendix~\ref{app:proof_query_update}. The multi-output covariance $\mathrm{Cov}_{\mathrm{comp}}\cc{f(x), \Delta}$ generalizes the scalar covariance in \citet{wu2026knowledge}; setting $m{=}1$, $U$ to the identity, and $2\sigma_{\mathrm{comp}}^2 = 1$ recovers their result.

As in the case of VoI for the direct observations, the inner maximum is handled by the one-shot reformulation with $K$ fantasy evaluation points \citep{balandat2020botorch,wu2020practical}:
\begin{equation*}
    \widehat{\alpha}_{\mathrm{comp}}(x_a, x_b, x'_1, \dots, x'_K; \mathcal{D}) = \sum_{d \in \set{0,1}} p(d \mid x_a, x_b, \mathcal{D})\, \max_{k} \E_{\mathcal{D}'}\!\cc{U(f(x'_k))} - u(\mathcal{D}).
\end{equation*}
Both $\widehat{\alpha}_{\mathrm{eval}}$ and $\widehat{\alpha}_{\mathrm{comp}}$ are optimized with Adam, gradient clipping, box projection onto $[0,1]^d$, and Sobol multi-start initialization.

\section{Experimental results}
\label{sec:experiments}

\begin{figure}[t]
\centering
\includegraphics[width=\linewidth]{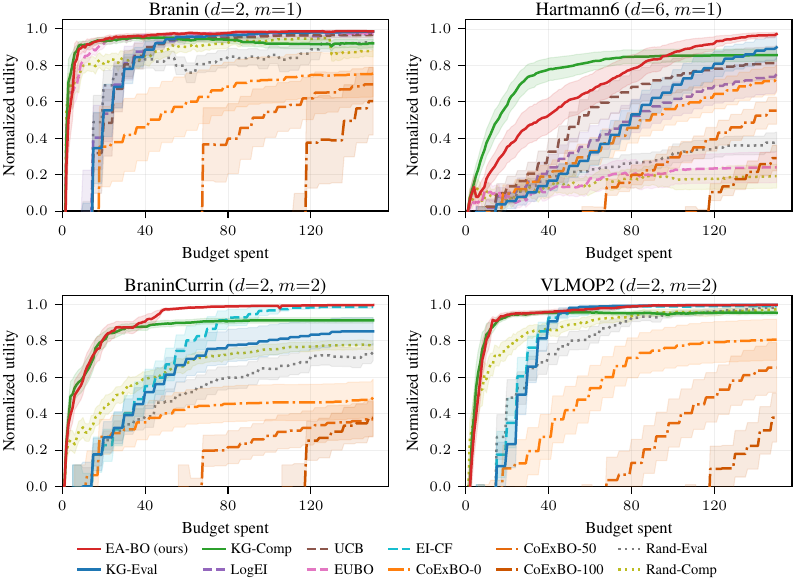}
\caption{Convergence trajectories under linear utility on four benchmark problems. Each panel plots normalized utility of the current best recommendation against cumulative budget spent ($c_{\mathrm{eval}}{=}5$, $c_{\mathrm{comp}}{=}1$, budget${=}150$). Lines represent averages over 50 repetitions; shaded regions are 95\% confidence intervals. \textsc{EA-BO} matches the fast early convergence of the comparison-only KG-Comp while reaching final utility comparable to or higher than all baselines, including the evaluation-only KG-Eval, and consistently outperforms CoExBO, the closest alternative from the HITL BO literature.}
\label{fig:trajectories}
\end{figure}

\textbf{Test functions.} We evaluate the proposed method on four standard single- and multi-objective benchmark functions: Branin ($d{=}2$, $m{=}1$), Hartmann6 ($d{=}6$, $m{=}1$), BraninCurrin ($d{=}2$, $m{=}2$), and VLMOP2 ($d{=}2$, $m{=}2$). All inputs lie in $[0,1]^d$ and each output component is standardized to zero mean and unit variance. We consider two utility functions: linear utility $U(y) = w^\top y$ with equal weights, and Chebyshev utility $U(y; w) = \min_j w_j y_j$, which is non-smooth and tests robustness of the acquisition under nonlinear scalarization (see Appendix~\ref{app:nonlinear}). We report normalized utility $U(f(\hat{x}_t)) / U(f(x^*))$.

\textbf{Cost model and noise levels.} Each trial starts with no initial data and runs under a total budget of $\Lambda = 150$ cost units, with direct evaluation cost $c_{\mathrm{eval}} = 5$ and pairwise comparison cost $c_{\mathrm{comp}} = 1$, giving a five-to-one cost ratio that models the plausible scenario where expert comparisons are substantially cheaper than physical experiments. Observation noise is $\sigma_{\mathrm{eval}} = 0.1$ (diagonal, equal across outputs) and comparison noise is $\sigma_{\mathrm{comp}} = 0.1$. All results are averaged over 50 repetitions.

\textbf{Baselines.} We compare \textsc{EA-BO} against single-source baselines, all using the same GP surrogate. The evaluation-only baselines are LogEI \citep{ament2023unexpected} and UCB \citep{srinivas2010gaussian} (single-objective), EI-CF \citep{astudillo2019composite} (multi-objective) and the comparison-only baseline is EUBO \citep{astudillo2023qeubo}. KG-Eval and KG-Comp are ablations of \textsc{EA-BO} restricted to a single source; Rand-Eval and Rand-Comp are uniform-random baselines. We also compare against CoExBO \citep{adachi2024looping}, a HITL BO method that trains an auxiliary preference model to guide candidate selection rather than incorporating comparisons as likelihood evidence in the surrogate. 

\subsection{Comparison to single-source BO and HITL BO methods}
\label{sec:results}

Figure~\ref{fig:trajectories} shows the optimization trajectories under linear utility as a function of cumulative budget spent. In the early phase, \textsc{EA-BO} allocates predominantly to cheap pairwise comparisons and, consequently, improves at the same rate as comparison-only KG-Comp. While the improvement of KG-Comp plateaus once binary signal saturates, \textsc{EA-BO} shifts toward direct evaluations as the posterior sharpens, and its utility continues to improve matching or exceeding evaluation-only methods. The first row of Table~\ref{tab:noise_allocation} shows the distribution of actions. The cost-aware selection rule combines the fast initial progress of comparison-based methods with the asymptotic accuracy of evaluation-based methods, without manual scheduling.

\textsc{EA-BO} achieves the best or statistically tied-best utility at budget exhaustion on all the single-objective tasks and multi-objective tasks with linear utility (See Table~\ref{tab:main_results}, Appendix~\ref{app:final_utility}.). The advantage is most pronounced on Hartmann6 ($d{=}6$), where the utility of cheap comparisons for early exploration is increased by a large search space.

Comparison-only methods (KG-Comp, EUBO) degrade in high dimensions: pairwise comparisons provide only binary signal. Thus the amount of queries required to approximate the posterior with a high accuracy increases significantly as $d$ grows. KG-Comp outperforms EUBO, which is consistent with the findings by \citet{wu2026knowledge}. Evaluation-only baselines (LogEI, UCB, EI-CF) perform well but cannot benefit from the cheaper pairwise comparison channel.

\begin{figure}[t]
\centering
\includegraphics[width=\linewidth]{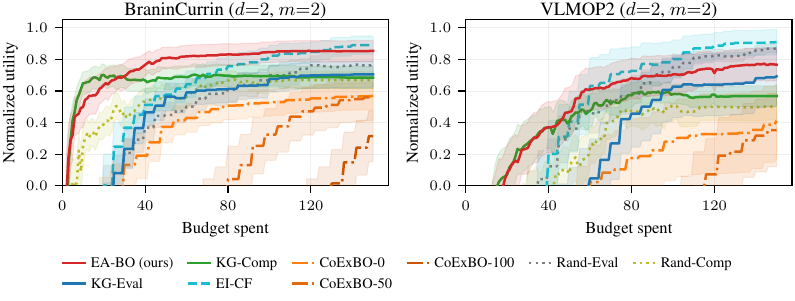}
\caption{Convergence trajectories under Chebyshev utility on two multi-objective benchmark problems. Each panel plots normalized utility of the current best recommendation against cumulative budget spent ($c_{\mathrm{eval}}{=}5$, $c_{\mathrm{comp}}{=}1$, budget${=}150$). Lines represent averages over 50 repetitions; shaded regions are 95\% confidence intervals. \textsc{EA-BO} matches the fast early convergence of the comparison-only KG-Comp and its utility at budget exhaustion is comparable to the utilities of the evaluation-only baselines.  \textsc{EA-BO} consistently outperforms the HITL BO baseline CoExBO.}
\label{fig:trajectories_cheb}
\end{figure}

\textbf{Chebyshev utility.} Under Chebyshev utility (Table~\ref{tab:chebyshev}, Appendix~\ref{app:final_utility}; Figure~\ref{fig:trajectories_cheb}), \textsc{EA-BO} again improves faster at the early stage the evaluation-only baselines. On BraninCurrin, \textsc{EA-BO} reaches the utility at budget exhaustion that is comparable to EI-CF while clearly outperforming both KG-Eval and KG-Comp. On VLMOP2, \textsc{EA-BO} improves faster than the evaluation-only baseline EI-CF in the early stage, but is overtaken by it at budget exhaustion. 

\textbf{Comparison to HITL BO.} \textsc{EA-BO} substantially outperforms all variants of CoExBO \citep{adachi2024looping} on all test functions (Figure~\ref{fig:trajectories}, Table~\ref{tab:main_results}). CoExBO optionally pretrains its preference model with 50 or 100 queries (CoExBO-50/100); pretraining improves the convergence rate but the pretraining cost spent cannot be recovered before budget exhaustion.

\subsection{Ablations}
\label{sec:ablation}

\textbf{Cost ratio sensitivity.} The utilities at budget exhaustion under different cost ratios $c_{\mathrm{eval}} / c_{\mathrm{comp}}$ on Hartmann6 are reported in Table~\ref{tab:cost_ablation}. \textsc{EA-BO} drops comparisons entirely when evaluations are cheaper than them ($c_{\mathrm{eval}}/c_{\mathrm{comp}} < 1$), and allocates 3--41\% of the budget to comparisons as the ratio grows. 

\begin{table}[t]
\small
\centering
\caption{Effect of cost ratio $c_{\mathrm{eval}} / c_{\mathrm{comp}}$ on comparison allocation and the utility at budget exhaustion on Hartmann6 (budget${=}150$, averages over 30 repetitions). C\% is the fraction of budget spent on comparisons. \textsc{EA-BO} abandons comparisons when evaluations are cheaper ($c_{\mathrm{eval}}/c_{\mathrm{comp}} < 1$), and increases allocation as the ratio grows.}
\label{tab:cost_ablation}
\begin{tabular}{ccccc}
\toprule
$c_\mathrm{eval}$ & $c_\mathrm{comp}$ & $c_\mathrm{eval} / c_\mathrm{comp}$ & C\% & \textsc{EA-BO} \\
\midrule
$1$ & $2$ & $1/2$ & 0\% & $0.983 \pm 0.019$ \\
\midrule
$1$ & $1$ & $1$ & 3\% & $0.988 \pm 0.017$ \\
\midrule
$2$ & $1$ & $2$ & 13\% & $0.954 \pm 0.193$ \\
$5$ & $1$ & $5$ & 30\% & $0.986 \pm 0.023$ \\
$10$ & $1$ & $10$ & 36\% & $0.924 \pm 0.137$ \\
$20$ & $1$ & $20$ & 41\% & $0.860 \pm 0.147$ \\
\bottomrule

\end{tabular}
\end{table}

\textbf{Noise sensitivity.} Table~\ref{tab:noise_allocation} and Figure~\ref{fig:ablation_noise} (Appendix~\ref{app:noise_sensitivity}) show the effect of varying observation and comparison noise on Hartmann6. \textsc{EA-BO} front-loads comparisons: early-phase allocation exceeds 90\% when comparisons are informative, dropping to ${\sim}14$\% late as evaluations become more valuable. Observation noise has little effect on allocation; final utility decreases monotonically with $\sigma_{\mathrm{eval}}$ as noisier evaluations make late-stage exploitation harder.

Under increasing $\sigma_{\mathrm{comp}}$, comparisons are progressively abandoned (31\% early at $\sigma_{\mathrm{comp}}{=}1.0$). Observe a non-monotonic dip at $\sigma_{\mathrm{comp}}{=}0.3$: the noise degrades comparisons but the myopic VoI still assigns to them a positive value mid-run. At higher $\sigma_{\mathrm{comp}}$, VoI abandons comparisons sooner and, consequently, the utility recovers.

\begin{table}[t]
\small
\centering
\caption{Effect of noise on comparison allocation and utility at budget exhaustion on Hartmann6 ($c_{\mathrm{eval}}{=}5$, $c_{\mathrm{comp}}{=}1$, 30 seeds). C\% early/late are comparison fractions in the first/last budget quartile. \textsc{EA-BO} front-loads comparisons when informative and shifts toward evaluations as the posterior sharpens.}
\label{tab:noise_allocation}
\begin{subtable}[t]{0.48\linewidth}
\centering
\subcaption{Observation noise $\sigma_e$}
\label{tab:noise_obs}
\begin{tabular}{cccc}
\toprule
$\sigma_\mathrm{eval}$ & C\% early & C\% late & Utility \\
\midrule
$0.1$ & 95\% & 4\% & $0.986 \pm 0.023$ \\
$0.2$ & 96\% & 0\% & $0.951 \pm 0.141$ \\
$0.3$ & 93\% & 1\% & $0.920 \pm 0.186$ \\
$0.5$ & 97\% & 3\% & $0.909 \pm 0.158$ \\
$1.0$ & 97\% & 7\% & $0.880 \pm 0.153$ \\
\bottomrule

\end{tabular}
\end{subtable}
\hfill
\begin{subtable}[t]{0.48\linewidth}
\centering
\subcaption{Comparison noise $\sigma_c$}
\label{tab:noise_comp}
\begin{tabular}{cccc}
\toprule
$\sigma_\mathrm{comp}$ & C\% early & C\% late & Utility \\
\midrule
$0.1$ & 95\% & 4\% & $0.986 \pm 0.023$ \\
$0.2$ & 92\% & 3\% & $0.941 \pm 0.130$ \\
$0.3$ & 85\% & 14\% & $0.795 \pm 0.343$ \\
$0.5$ & 87\% & 10\% & $0.909 \pm 0.211$ \\
$1.0$ & 31\% & 2\% & $0.853 \pm 0.251$ \\
\bottomrule

\end{tabular}
\end{subtable}
\end{table}

\section{Discussion}

We presented a GP-based BO method that treats both pairwise comparisons and direct evaluations as noisy evidence about the objective, with a cost-aware VoI acquisition function that adaptively allocates budget between the two sources. The method matches or exceeds single-source strategies on all benchmarks, exploiting cheap comparisons when informative and falling back to evaluations when more accuracy is required, and outperforms existing methods that also combine both information sources but use comparisons indirectly.

\textbf{Limitations and future work.}
Our model assumes that the expert's comparative judgements are governed by $U \circ f$. In practice, the expert may rely on a mental model $g: \mathcal{X} \rightarrow \mathbb{R}^m$ that deviates from $f$. Currently, these deviations are interpreted as observation noise; in future work we will consider a setting where the correlation structure between $f$ and $g$ is known or learnable. In the multi-objective case, we assume that the utility function $U$ is known; it may, for instance, have been elicited before the optimization process, as in \emph{a priori} multi-objective optimization methods. We will consider extension to interactive multi-objective optimization, where $U$ is also learned during the optimization process, in future work.

\bibliography{knowledge_moo}

\appendix

\section{Proof of the query posterior mean update}
\label{app:proof_query_update}

We prove the identity stated in Section~\ref{sec:VoI_pairwise_comparisons} for the fantasy posterior mean after a pairwise comparison, adapting Lemma 3.1 of \citet{wu2026knowledge} to expert noise $\sigma_{\mathrm{comp}}$ and linear utility $U$.

\subsection{Setup}

Under the variational posterior, the triple $(f(x), f(x_a), f(x_b))$ is jointly Gaussian with means $(\mu(x), \mu(x_a), \mu(x_b))$ and covariances inherited from the posterior kernel $K(\cdot, \cdot)$. Let $U: \R^m \to \R$ be linear, $U(y) = w^\top y$ for some $w \in \R^m$. Define the utility difference
\begin{equation*}
    \Delta := U(f(x_a)) - U(f(x_b)) = w^\top (f(x_a) - f(x_b)).
\end{equation*}
Under the variational posterior $\Delta$ is Gaussian with moments
\begin{align*}
    \mu_\Delta &= w^\top (\mu(x_a) - \mu(x_b)), \\
    \sigma_\Delta^2 &= w^\top \left[ K(x_a, x_a) + K(x_b, x_b) - 2 K(x_a, x_b) \right] w,
\end{align*}
and covariance with $f(x)$
\begin{equation*}
    \gamma := \mathrm{Cov}\cc{f(x), \Delta} = K(x, x_a) w - K(x, x_b) w \in \R^m.
\end{equation*}
The comparison likelihood of Section~\ref{sec:surrogate} encodes the event $\set{d = 1}$ as $\set{\Delta + \varepsilon_{\mathrm{comp}} > 0}$ for $\varepsilon_{\mathrm{comp}} \sim N(0, 2\sigma_{\mathrm{comp}}^2)$ independent of $f$. Writing
\begin{equation*}
    \nu^2 := \sigma_\Delta^2 + 2\sigma_{\mathrm{comp}}^2, \qquad \tau := \frac{\mu_\Delta}{\nu},
\end{equation*}
the marginal probability of the event is $\Pr(d = 1) = \Phi(\tau)$ by the standard Gaussian tail identity. We claim
\begin{equation}
\label{eq:posterior_mean_update}
    \E\cc{f(x) \mid d = 1} = \mu(x) + \frac{\phi(\tau)}{\Phi(\tau)} \cdot \frac{\gamma}{\nu},
\end{equation}
which is the formula used in Section~\ref{sec:VoI_pairwise_comparisons}.

\subsection{Proof}

Introduce the auxiliary variable $T := \Delta + \varepsilon_{\mathrm{comp}}$. Since $\varepsilon_{\mathrm{comp}}$ is independent of $f$ and Gaussian, the pair $(f(x), T)$ is jointly Gaussian with
\begin{equation*}
    \E\cc{T} = \mu_\Delta, \qquad \mathrm{Var}\cc{T} = \sigma_\Delta^2 + 2\sigma_{\mathrm{comp}}^2 = \nu^2, \qquad \mathrm{Cov}\cc{f(x), T} = \gamma.
\end{equation*}
The event of interest is $\set{T > 0}$, a half-space constraint on the auxiliary Gaussian.

We use the classical identity for the mean of a jointly Gaussian random variable under a half-space constraint \citep{azzalini2013skew}.
If $(X, T)$ is jointly Gaussian with $\E\cc{X} = \mu_X$, $\E\cc{T} = \mu_T$, $\mathrm{Var}\cc{T} = \sigma_T^2$, and $\mathrm{Cov}\cc{X, T} = \gamma_X$, then
\begin{equation*}
    \E\cc{X \mid T > 0} = \mu_X + \frac{\gamma_X}{\sigma_T} \cdot \frac{\phi(\mu_T / \sigma_T)}{\Phi(\mu_T / \sigma_T)}.
\end{equation*}
The identity applies componentwise when $X$ is vector-valued, with $\gamma_X$ a vector of covariances. Applying it with $X = f(x)$, $\mu_X = \mu(x)$, $\mu_T = \mu_\Delta$, $\sigma_T = \nu$, and $\gamma_X = \gamma$ yields
\begin{equation*}
    \E\cc{f(x) \mid d = 1} = \mu(x) + \frac{\gamma}{\nu} \cdot \frac{\phi(\tau)}{\Phi(\tau)},
\end{equation*}
which is \eqref{eq:posterior_mean_update}. The case $d = 0$ follows by replacing the event $\set{T > 0}$ with $\set{T < 0}$, equivalently by replacing $\Delta$ with $-\Delta$, and gives the same form with $\tau$ replaced by $-\tau$.

\section{Computational details}
\label{app:elbo_details}
\label{app:nonlinear}

This appendix collects the closed-form expressions and approximations used in the ELBO (Section~\ref{sec:vi}) and VoI acquisition (Sections~\ref{sec:VoI_observations} and \ref{sec:VoI_pairwise_comparisons}). Each quantity is presented first for linear $U$, where $U$ commutes with Gaussian expectations, then for nonlinear $U$ such as Chebyshev or $L^2$ scalarizations.

We begin with the ELBO derivation. The marginal log-likelihood $\log p(\mathcal{D} \mid \theta)$ is intractable due to the probit factors. Writing it as an expectation under the joint $p(f, u \mid \theta)$ and applying Jensen's inequality gives
\begin{equation*}
    \log p(\mathcal{D} \mid \theta) \geq \E_{q(f, u)} \cc{\log \frac{p(f, u \mid \theta)\, p(\mathcal{D} \mid f)}{q(f, u)}} =: \mathcal{L}(q, \theta).
\end{equation*}
Substituting the SVGP factorization $q(f, u) = p(f \mid u)\, q(u)$ cancels the $p(f \mid u)$ terms in numerator and denominator, and the dataset factorization $p(\mathcal{D} \mid f) = \prod_i p(y_i \mid f) \prod_j p(d_j \mid f)$ yields the decomposed form of the main text.

\subsection{ELBO terms}

\paragraph{Observation term.}
Each observation term is a Gaussian expectation of a Gaussian log density,
\begin{equation*}\begin{split}
    \E_{z \sim N(\mu_{\mathrm{comp}}, \Sigma_{\mathrm{comp}})}\!\cc{\log N(z \mid \mu_p, \Sigma_p)}
    =& -\tfrac{1}{2}\!\big[ m \log 2\pi + \log |\Sigma_p| + \mathrm{tr}(\Sigma_p^{-1} \Sigma_{\mathrm{comp}}) \\ 
    &+ (\mu_{\mathrm{comp}} - \mu_p)^\top \Sigma_p^{-1} (\mu_{\mathrm{comp}} - \mu_p) \big],
\end{split}
\end{equation*}
evaluated at $\mu_{\mathrm{comp}} = \mu_{\mathrm{comp}}(x_i)$, $\Sigma_{\mathrm{comp}} = \Sigma_{\mathrm{comp}}(x_i, x_i)$, $\mu_p = y_i$, and $\Sigma_p = \Sigma_{\mathrm{eval}}$.

\paragraph{Query term, linear $U$.}
For linear $U(y) = w^\top y$, the utility difference $\Delta_j = U(f(x_j)) - U(f(x'_j))$ is Gaussian under $q$ with mean $\mu_{\Delta_j} = w^\top(\mu(x_j) - \mu(x'_j))$ and variance $\sigma_{\Delta_j}^2 = w^\top[K(x_j,x_j) + K(x'_j,x'_j) - 2K(x_j,x'_j)]w$. The expected log-likelihood reduces to a one-dimensional integral,
\begin{equation*}
    \E_{\mathrm{comp}}\!\cc{\log p(d_j \mid x_j, x'_j, f)} = \E_{\Delta \sim N(\mu_{\Delta_j},\, \sigma_{\Delta_j}^2)}\!\cc{\log \Phi\!\left(\frac{(2d_j - 1)\,\Delta}{\sqrt{2}\,\sigma_{\mathrm{comp}}}\right)},
\end{equation*}
computed by one-dimensional Gauss--Hermite quadrature.

\paragraph{Query term, nonlinear $U$.}
When $U$ is nonlinear, $\Delta_j$ is not Gaussian under $q$. We approximate its distribution by a Gaussian matched to its first two moments,
\begin{align*}
    \E_{\mathrm{comp}}[\Delta_j] &= \E_{\mathrm{comp}}[U(f(x_j))] - \E_{\mathrm{comp}}[U(f(x'_j))], \\
    \mathrm{Var}_{\mathrm{comp}}[\Delta_j] &= \mathrm{Var}_{\mathrm{comp}}[U(f(x_j))] + \mathrm{Var}_{\mathrm{comp}}[U(f(x'_j))] - 2\,\mathrm{Cov}_{\mathrm{comp}}[U(f(x_j)), U(f(x'_j))],
\end{align*}
where each term is computed by $m$-dimensional Gauss--Hermite quadrature on the marginals $q(f(x_j))$ and $q(f(x'_j))$. The covariance term requires the joint $q(f(x_j), f(x'_j))$ and is evaluated by $2m$-dimensional quadrature, which is tractable for moderate $m$. The one-dimensional rule for $\E[\log\Phi(\cdot)]$ is then applied to the matched Gaussian, as in the linear case.

\paragraph{KL term.}
The KL divergence between $q(u) = N(m_u, S_u)$ and $p(u \mid \theta) = N(0, K_{ZZ})$ is
\begin{equation*}
    \mathrm{KL}(q(u) \,\|\, p(u)) = \tfrac{1}{2}\!\left[ \mathrm{tr}(K_{ZZ}^{-1} S_u) + m_u^\top K_{ZZ}^{-1} m_u - M + \log |K_{ZZ}| - \log |S_u| \right].
\end{equation*}

\subsection{Fantasy posterior updates}
\label{app:fantasy_updates}

\paragraph{Sample fantasy mean.}
The fantasy posterior mean after a Gaussian observation at $x_\mathrm{cand}$ is
\begin{equation*}
    \mu_{\mathcal{D}'}(x) = \mu(x) + K(x, x_\mathrm{cand}) \left[ K(x_\mathrm{cand}, x_\mathrm{cand}) + \Sigma_{\mathrm{eval}} \right]^{-1} (y - \mu(x_\mathrm{cand})).
\end{equation*}
This is exact and independent of $U$.

\paragraph{Query fantasy mean, linear $U$.}
For linear $U$ the closed-form update is derived in Appendix~\ref{app:proof_query_update}:
\begin{equation*}
    \mu_{\mathcal{D} \cup \{((x_a,x_b),1)\}}(x) = \mu(x) + \frac{\phi(\tau)}{\Phi(\tau)} \cdot \frac{\gamma}{\nu},
\end{equation*}
where $\gamma = K(x,x_a)w - K(x,x_b)w$, $\nu = \sqrt{\sigma_\Delta^2 + 2\sigma_{\mathrm{comp}}^2}$, and $\tau = \mu_\Delta / \nu$.

\paragraph{Query fantasy mean, nonlinear $U$.}
When $U$ is nonlinear, $\Delta$ is not Gaussian and the closed-form update does not apply directly. We approximate $\Delta$ by the moment-matched Gaussian described above and apply the same formula with the matched moments substituted for $\mu_\Delta$ and $\sigma_\Delta^2$. The result is exact for linear $U$ and approximate otherwise, with the approximation error controlled by the curvature of $U$ and the posterior variance of $f$ at $x_a$ and $x_b$.

\subsection{Dataset utility and VoI evaluation}
\label{app:dataset_utility}

\paragraph{Dataset utility.}
For linear $U$ the expectation and $U$ commute, giving $u(\mathcal{D}) = \max_x U(\mu_\mathcal{D}(x))$. For nonlinear $U$ we compute $\E_f[U(f(x)) \mid \mathcal{D}]$ by $m$-dimensional Gauss--Hermite quadrature on $q(f(x))$. The same applies to the final recommendation $\hat{x}$.

\paragraph{Sample VoI.}
For linear $U$, $U(\mu_{\mathcal{D}'}(x))$ is linear in $y$ and the sample VoI reduces to the knowledge gradient \citep{frazier2009knowledge,wu2016parallel}. For nonlinear $U$, $U(\mu_{\mathcal{D}'}(x))$ is no longer linear in $y$; we draw fantasy observations using Matheron's rule for pathwise conditioning \citep{wilson2020efficiently,wilson2021pathwise} and average $\max_k U(\mu_{\mathcal{D}'}(x'_k))$ over the fantasy samples.

\paragraph{Query VoI.}
The query VoI is evaluated by the two-term discrete sum of Section~\ref{sec:VoI_pairwise_comparisons} using the fantasy means above; no further approximation is needed beyond the moment matching in the nonlinear case.

\section{Extended experimental results}
\label{app:extended_results}

\subsection{Final utility at budget exhaustion}
\label{app:final_utility}
See Table~\ref{tab:main_results} and \ref{tab:chebyshev} for final utility values of experimental results with linear utility and Chebyshev utility, respectively.

\begin{table}[h]
\small
\centering
\caption{Normalized utility at budget exhaustion (mean $\pm$ std over 50 seeds). Best result per problem in \textbf{bold}; results within one standard error of the best are underlined. Methods marked ``---'' are not applicable to that problem class.}
\label{tab:main_results}
\begin{tabular}{lcccc}
\toprule
Method & Branin & Hartmann6 & BraninCurrin & VLMOP2 \\
\midrule
\textsc{EA-BO (ours)} & $\underline{0.988 \pm 0.012}$ & $\mathbf{0.984 \pm 0.023}$ & $\mathbf{0.999 \pm 0.003}$ & $\mathbf{0.998 \pm 0.002}$ \\
\textsc{KG-Eval} & $\mathbf{0.988 \pm 0.014}$ & $0.929 \pm 0.220$ & $0.853 \pm 0.273$ & $\underline{0.998 \pm 0.002}$ \\
\textsc{KG-Comp} & $0.921 \pm 0.086$ & $0.874 \pm 0.112$ & $0.912 \pm 0.054$ & $0.955 \pm 0.044$ \\
\midrule
\textsc{LogEI} & $0.978 \pm 0.019$ & $0.810 \pm 0.319$ & {---} & {---} \\
\textsc{UCB} & $0.973 \pm 0.051$ & $0.847 \pm 0.328$ & {---} & {---} \\
\textsc{EUBO} & $0.977 \pm 0.036$ & $0.211 \pm 0.269$ & {---} & {---} \\
\textsc{EI-CF} & {---} & {---} & $0.989 \pm 0.025$ & $0.992 \pm 0.009$ \\
\midrule
\textsc{CoExBO-0} & $0.767 \pm 0.452$ & $0.813 \pm 0.241$ & $0.514 \pm 0.370$ & $0.836 \pm 0.377$ \\
\textsc{CoExBO-50} & $0.736 \pm 0.316$ & $0.807 \pm 0.180$ & $0.409 \pm 0.373$ & $0.753 \pm 0.431$ \\
\textsc{CoExBO-100} & $0.706 \pm 0.656$ & $0.609 \pm 0.279$ & $0.517 \pm 0.349$ & $0.632 \pm 0.517$ \\
\midrule
\textsc{Rand-Eval} & $0.950 \pm 0.064$ & $0.464 \pm 0.218$ & $0.818 \pm 0.154$ & $0.994 \pm 0.009$ \\
\textsc{Rand-Comp} & $0.887 \pm 0.095$ & $0.143 \pm 0.174$ & $0.808 \pm 0.074$ & $0.980 \pm 0.028$ \\
\bottomrule

\end{tabular}
\end{table}

\begin{table}[h]
\small
\centering
\caption{Normalized utility at budget exhaustion under Chebyshev utility (mean $\pm$ std). Format follows Table~\ref{tab:main_results}.}
\label{tab:chebyshev}
\begin{tabular}{lcc}
\toprule
Method & BraninCurrin & VLMOP2 \\
\midrule
\textsc{EA-BO (ours)} & $0.863 \pm 0.233$ & $0.782 \pm 0.371$ \\
\textsc{KG-Eval} & $0.710 \pm 0.299$ & $\underline{0.911 \pm 0.318}$ \\
\textsc{KG-Comp} & $0.685 \pm 0.252$ & $0.571 \pm 0.228$ \\
\midrule
\textsc{EI-CF} & $\mathbf{0.915 \pm 0.190}$ & $\mathbf{0.915 \pm 0.284}$ \\
\midrule
\textsc{CoExBO-0} & $0.570 \pm 0.309$ & $0.412 \pm 0.862$ \\
\textsc{CoExBO-50} & $0.598 \pm 0.498$ & $0.399 \pm 0.830$ \\
\textsc{CoExBO-100} & $0.456 \pm 0.474$ & $-0.215 \pm 0.770$ \\
\midrule
\textsc{Rand-Eval} & $\underline{0.890 \pm 0.120}$ & $\underline{0.911 \pm 0.071}$ \\
\textsc{Rand-Comp} & $0.669 \pm 0.192$ & $0.498 \pm 0.305$ \\
\bottomrule

\end{tabular}
\end{table}

\subsection{Noise sensitivity}
\label{app:noise_sensitivity}
\begin{figure}[h]
\centering
\includegraphics[width=0.6\linewidth]{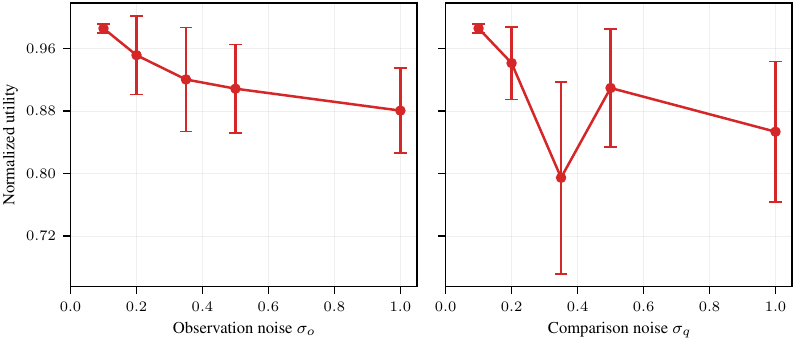}
\caption{Sensitivity of \textsc{EA-BO} to observation noise $\sigma_{\mathrm{eval}}$ (left) and comparison noise $\sigma_{\mathrm{comp}}$ (right) on Hartmann6 ($c_{\mathrm{eval}}{=}5$, $c_{\mathrm{comp}}{=}1$, 30 seeds).}
\label{fig:ablation_noise}
\end{figure}

See Figure~\ref{fig:ablation_noise} for the trajectory of final utility depending on comparison noise.

\subsection{Computational resources}

All experiments were run on an internal SLURM cluster. Each job was allocated a single NVIDIA V100 GPU (32\,GB), 2 CPU cores, and 8\,GB of RAM. The experimental grid comprised 3{,}800 jobs, of which 3{,}770 completed successfully and 30 timed out at the 8-hour SLURM limit. Individual runs took a median of 0.9 hours (mean 1.2 hours), for a total of approximately 4{,}500 GPU-hours. Preliminary experiments during method development required an estimated additional 500 GPU-hours beyond the reported results.

\end{document}